\documentclass[10pt,twocolumn,letterpaper]{article}

\usepackage{cvpr}
\usepackage{times}
\usepackage{epsfig}
\usepackage{graphicx}
\usepackage{amsmath}
\usepackage{amssymb}
\usepackage{fixltx2e}
\usepackage{caption}
\usepackage{subcaption}

\usepackage{rotating}
\usepackage{booktabs} \newcommand{\ra}[1]{\renewcommand{\arraystretch}{#1}}

\newcommand\blfootnote[1]{%
  \begingroup
  \renewcommand\thefootnote{}\footnote{#1}%
  \addtocounter{footnote}{-1}%
  \endgroup
}
\usepackage[pagebackref=true,breaklinks=true,letterpaper=true,colorlinks,bookmarks=false]{hyperref}

\cvprfinalcopy % *** Uncomment this line for the final submission

\def\cvprPaperID{1828} % *** Enter the CVPR Paper ID here

\ifcvprfinal\fi
\begin{document}

%%%%%%%%%%%%%%%%%%%%%%%%%%%%%%%%%%%%%%%%%%%%%%%%%%%%%%%%%%%%%%%
% -- TITLE
%%%%%%%%%%%%%%%%%%%%%%%%%%%%%%%%%%%%%%%%%%%%%%%%%%%%%%%%%%%%%%%
\title{From Image-level to Pixel-level Labeling with Convolutional Networks}

\author{Pedro O. Pinheiro$^{1,2}$\;\;\;\;\;\;\;\;\;\;\; Ronan Collobert$^{1,3,\dagger}$ \\
$^1$Idiap Research Institute, Martigny, Switzerland \\ 
$^2$Ecole Polytechnique F\'ed\'erale de Lausanne (EPFL), Lausanne, Switzerland\\
$^3$Facebook AI Research, Menlo Park, CA, USA\\
{\small \tt pedro@opinheiro.com\;\;\;\;\;\;\;\;\;\;\; ronan@collobert.com}
}

\maketitle
\blfootnote{\mbox{}$^{\dagger}$All research was conducted at the Idiap Research Institute, before Ronan Collobert joined Facebook AI Research.}
%\thispagestyle{empty}

%%%%%%%%%%%%%%%%%%%%%%%%%%%%%%%%%%%%%%%%%%%%%%%%%%%%%%%%%%%%%%%
% -- Abstract
%%%%%%%%%%%%%%%%%%%%%%%%%%%%%%%%%%%%%%%%%%%%%%%%%%%%%%%%%%%%%%%
\begin{abstract}

We are interested in inferring object segmentation by leveraging only
object class information, and by considering only minimal priors on the
object segmentation task. This problem could be viewed as a kind of weakly
supervised segmentation task, and naturally fits the Multiple Instance
Learning (MIL) framework: every training image is known to have (or not) at
least one pixel corresponding to the image class label, and the
segmentation task can be rewritten as inferring the pixels belonging to the
class of the object (given one image, and its object class).  We propose a
Convolutional Neural Network-based model, which is constrained during
training to put more weight on pixels which are important for classifying
the image. We show that at test time, the model has learned to discriminate
the right pixels well enough, such that it performs very well on an
existing segmentation benchmark, by adding only few smoothing priors.
Our system is trained using a subset of the Imagenet dataset and the
segmentation experiments are performed on the challenging Pascal VOC
dataset (with \emph{no} fine-tuning of the model on Pascal VOC). Our model
beats the state of the art results in weakly supervised object segmentation
task by a large margin. We also compare the performance of our model with
state of the art fully-supervised segmentation approaches.

\end{abstract}

%%%%%%%%%%%%%%%%%%%%%%%%%%%%%%%%%%%%%%%%%%%%%%%%%%%%%%%%%%%%%%%
% -- Introduction
%%%%%%%%%%%%%%%%%%%%%%%%%%%%%%%%%%%%%%%%%%%%%%%%%%%%%%%%%%%%%%%
\section{Introduction}
\label{sec-introduction}

Object segmentation is a computer vision tasks which consists in assigning
an object class to sets of pixels in an image. This task is extremely
challenging, as each object in the world generates an infinite number of
images with variations in position, pose, lightning, texture, geometrical
form and background. Natural image segmentation systems have to cope with
these variations, while being limited in the amount of available training
data. Increasing computing power, and recent releases of reasonably large segmentation datasets such as Pascal VOC~\cite{Everingham10} have nevertheless made the segmentation task a reality.

\begin{figure}[!t]
\begin{center}
   \includegraphics[width=\linewidth]{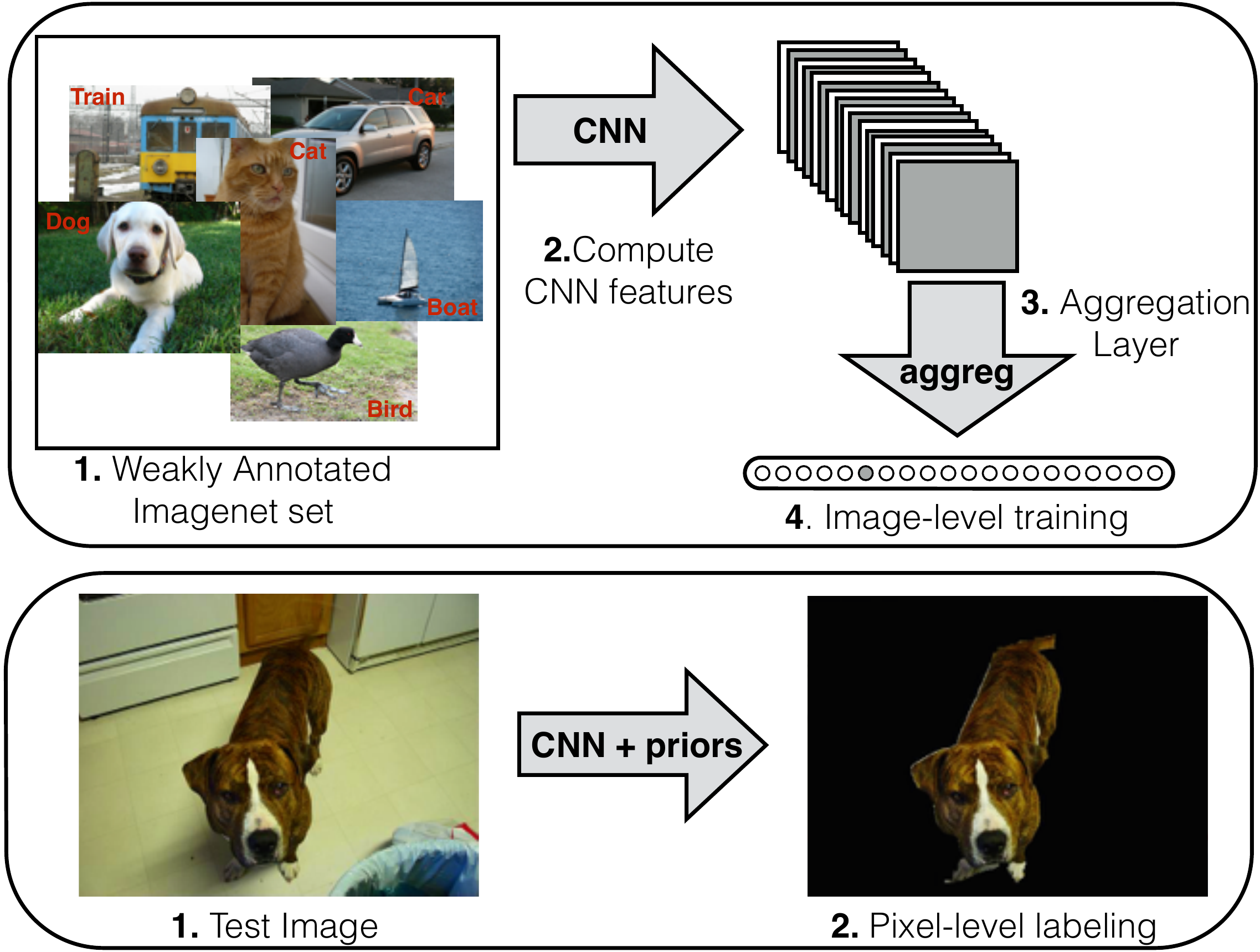}
\end{center}
   \caption{{\bf A schematic illustration of our method}. \emph{Top}:
     (1)~The model is trained using weakly annotated data (only image-level
     class information) from Imagenet.  (2)~The CNN generates feature planes. (3)~These planes pass through an aggregation layer to constrain the model to put more weight on the right pixels. (4)~The system is trained by classifying the correct image-level label. 
     \emph{Bottom}: During test time, the aggregation layer
     is removed and the CNN densely classifies every pixel of the image
     (considering only few segmentation priors).}
\label{fig:schematic}
\end{figure}

We rely on Convolutional Neural Networks (CNNs) ~\cite{lecun1998gradient},
an important class of algorithms which have been shown to be
state-of-the-art on large object recognition
tasks~\cite{NIPS2012_0534,GoogLenet}, as well as on fully supervised
segmentation task~\cite{Farabet:2012}. One advantage of CNNs is that they learn
sufficiently general features, and therefore they can excel in transfer learning:
\eg CNN models trained on the Imagenet classification database~\cite{imagenet_cvpr09} could be exploited for different vision tasks~\cite{girshick2014rcnn,hariharan14sds,RazavianASC14}.
Their main disadvantage, however, is the need of a large number of
fully-labeled dataset for training. 
Given that classification labels are much more abundant than segmentation labels, it is natural to find a bridge between classification and segmentation, which would transfer efficiently learned features from one task to the other one.
% Given that segmentation labels
% are more expensive than classification labels, it is natural to find
% a bridge between classification and segmentation, which would transfer
% efficiently learned features from one task to the other one.

Our CNN-based model is not trained with segmentation labels, nor bounding
box annotations. Instead, we only consider a single object class label for
a given image, and the model is constrained to put more weight on important
pixels for classification. This approach can be seen as an instance of
Multiple Instance Learning (MIL)~\cite{Maron98}. In this context, every
image is known to have (or not) -- through the image class label -- one or
several pixels matching the class label. However, the positions of these
pixels are unknown, and have to be inferred.

Because of computing power limitations, we built our model over the
\verb+Overfeat+ feature extractor, developed by Sermanet
\etal~\cite{Sermanet:2013vi}. This feature extractor correspond to the first layers of a CNN, well-trained over ImageNet. Features are fed into few extra convolutional
layers, which forms our ``segmentation network''.

Training is achieved by maximizing the classification likelihood over the
classification training set (subset of Imagenet), by adding an extra layer
to our network, which constrains the model to put more weight on pixels
which are important for the classification decision. At test time, the
constraining layer is removed, and the label of each image pixel is
efficiently inferred. Figure~\ref{fig:schematic} shows a general
illustration of our approach.

The paper is organized as follows. Section~\ref{related-works} presents
related work. Section~\ref{architecture} describes our architecture
choices. Section~\ref{experiments} compares our model with both weakly and
fully supervised state-of-the-art approaches. We conclude in
Section~\ref{conclusion}.

%%%%%%%%%%%%%%%%%%%%%%%%%%%%%%%%%%%%%%%%%%%%%%%%%%%%%%%%%%%%%%%
% -- Related Works
%%%%%%%%%%%%%%%%%%%%%%%%%%%%%%%%%%%%%%%%%%%%%%%%%%%%%%%%%%%%%%%
\section{Related Work}
\label{related-works}
Labeling data for segmentation task is difficult if compared to labeling data
for classification. For this reason, several weakly supervised object
segmentation systems have been proposed in the past few years.  
For instance, Vezhnevets and Buhmann~\cite{VezhnevetsB10} proposed an approach
based on Semantic Texton Forest, derived in the context of MIL. However, the model fails to model relationship between superpixels. To model these relationships,~\cite{eth_biwi_00866} introduced a graphical model -- named Multi-Image Model (MIM) -- to connect superpixels from all training images, based on their appearance similarity. The unary potentials of the MIM are initialized with the output of~\cite{VezhnevetsB10}.

In~\cite{VezhnevetsB12}, the authors define a parametric family of structured models, where each model carries visual cues in a different way. A maximum expected agreement model selection principle evaluates the quality of a model from a family. An algorithm based on Gaussian processes is proposed to efficiency search the best model for different visual cues.

More recently,~\cite{ZhangGXLSJ14} proposed an algorithm that learns the
distribution of spatially structural superpixel sets from image-level
labels. This is achieved by first extracting graphlets (small graphs consisting of superpixels and encapsulating their spatial structure) from a given
image.
%JEDI: this is not clear either
Labels from the training images are transfered into graphlets 
throughout a proposed manifold embedding algorithm. A Gaussian mixture model is then
used to learn the distribution of the post-embedding graphlets, \ie vectors output from the graphlet embedding. The inference is done by leveraging the learned GMM prior to
measure the structure homogeneity of a test image.

In contrast with previous approaches for weakly supervised segmentation, we
avoid designing task-specific features for segmentation. Instead, a CNN
learns the features: the model is trained through a cost function which
casts the problem of segmentation into the problem of finding pixel-level
labels from image-level labels. As we will see in Section~\ref{experiments}, learning the right features for segmentation leads to better performance compared to existing weakly supervised segmentation system. Another difference from our approach is that we train our model in a different dataset (Imagenet) from the one we validate the results (Pascal VOC).

\paragraph{Transfer Learning and CNNs} In the last few years, convolutional
networks have been widely used in the context of object recognition. A
notable system is the one from Krizhevsky \etal~\cite{NIPS2012_0534}, which
performs very well on Imagenet. In~\cite{Oquab14} the authors built upon
Krizhevsky's approach and showed that a model trained for classification on
Imagenet dataset can be used for classification in a different dataset
(namely Pascal VOC) by taking into account the bounding box information.
In a recent yet unpublished work~\cite{Oquab13}, the authors adapt an
Imagenet-trained CNN to the Pascal VOC classification task. The network is
fine-tuned on Pascal VOC, by modifying the cost function to include a final
max-pooling layer. Similar to our aggregation layer, the max-pooling
outputs a single image-level score for each of the classes. In contrast,
(1)~we not limit ourselves to the Pascal VOC classification problem, but
tackle the more challenging problem of segmentation and (2)~our model is
not fine-tuned on Pascal VOC.

In the same spirit, Girshick \etal~\cite{girshick2014rcnn} showed that a
model trained for classification on Imagenet can be adapted for object
detection on Pascal VOC. The authors proposed to combine bottom-up techniques for
generating detection region candidates with CNNs. The authors
achieved state-of-the-art performance in object detection. Based upon this
work,~\cite{hariharan14sds} derived a model that detects all instances of a
category in an image and, for each instance, marks the pixels that belong
to it. Their model, entitled SDS (Simultaneous Detection and Segmentation),
uses category-specific, top-down figure-ground predictions to refine
bottom-up detection candidates. 

As for these existing state-of-the-art approaches, our system leverages
features learned over the Imagenet classification  dataset. However,
our approach differs from theirs in some important aspects. Compared to~\cite{girshick2014rcnn,Oquab14}, we consider the more challenging problem of object segmentation and do not use any information other than the image-level annotation.~\cite{Oquab13} consider a weakly supervised scenario, but only deals with the classification problem. Compared to~\cite{hariharan14sds}, we consider only the the image-level annotation to infer the pixel-level one.
In that respect, we do not use any segmentation information (our model is not refined over the segmentation data either), nor bounding box annotation during the training period. One could argue that a classification dataset like Imagenet has somewhat already cropped properly objects. While this might true for certain objects, it is not the case for many images, and in any case the ``bounding box'' remains quite loose.

\begin{figure*}[!t]
\begin{center}
   \includegraphics[width=\linewidth]{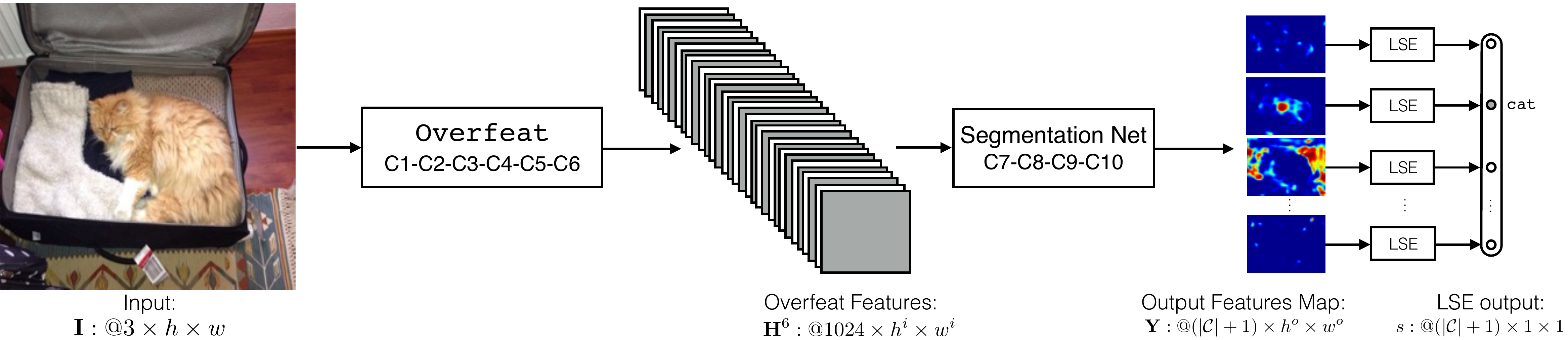}
\end{center}
   \caption{{\bf Outline of our architecture.} The full RGB image is
     forwarded through the network (composed of Overfeat and four extra
     convolutional features), generating output planes of dimension
     $(|\mathcal{C}|+1)\times h^o\times w^o$. These output planes can be seen as
     pixel-level labels of a sub-sampled version of the input image. The
     output then passes through the \emph{Log-Sum-Exp} layer to aggregate
     pixel-level labels into image-level ones. The error is backpropagated
     through layers C10-C7.}
\label{fig:train}
\end{figure*}

%%%%%%%%%%%%%%%%%%%%%%%%%%%%%%%%%%%%%%%%%%%%%%%%%%%%%%%%%%%%%%%
% -- Architecture
%%%%%%%%%%%%%%%%%%%%%%%%%%%%%%%%%%%%%%%%%%%%%%%%%%%%%%%%%%%%%%%
\section{Architecture}
\label{architecture}

As we pointed out in Section~\ref{sec-introduction}, CNNs are a very flexible model which can be applied on various image
processing tasks, as they alleviate the need of task-specific features. CNNs
learn a hierarchy of filters, which extract higher level of representations
as one goes ``deeper'' in the hierarchy~\cite{ZeilerECCV14}. The type of
features they learn is also sufficiently general that CNNs make transfer
learning (to another task) quite easy. The main drawback of these models,
however, is that a large amount of data is necessary during training.

Since the number of image-level object labels is much bigger than pixel-level segmentation labels, it is thus natural to leverage image classification datasets for performing segmentation. 
In the following, we consider a problem of segmentation with a set of classes~$\mathcal{C}$. We assume the classification dataset contains
\emph{at least} the same classes. Extra classes available at classification
time, but which are not in the segmentation dataset are mapped to a
``background'' class. This background class is essential to limit the
number of false positive during segmentation.

Our architecture is a CNN, which is trained over a subset of Imagenet, to produce pixel-level labels from image-level labels.
As shown in Figure~\ref{fig:train}, our CNN is quite standard, with 10 levels of convolutions and (optional) pooling. It takes as input a $400\times 400$
RGB patch $I$, and outputs $|\mathcal{C}|+1$ planes (one per class, plus the background
class) corresponding to the score of the 12-times downsampled image
\emph{pixels labels}. During training, an extra layer, described in
Section~\ref{dmil}, aggregates pixel-level labels into an image-level
label. For computational power reasons, we ``froze'' the first layers of our CNN,
to the ones of some already well-trained (over Imagenet classification
data) CNN model. 

We pick \verb+Overfeat+~\cite{Sermanet:2013vi}, trained to perform object classification on the ILSVRC13 challenge. The \verb+Overfeat+ model generates feature maps of dimensions $1024\times h^{i}\times w^{i}$, where $h^{i}$ and $w^{i}$ are functions of the size of the RGB input image, the convolution kernel sizes, convolution strides and max-pooling sizes. Keeping only the first 6 convolution layers and 2 pooling layers of \verb+Overfeat+, our RGB $400\times400$ image patch $I$ is transformed into a $1024\times29\times29$ feature representation.

We add four extra convolutional layers (we denote $\mathbf{H}^6$ for feature planes coming out from \verb+OverFeat+). Each of them (but the last one $\mathbf{Y}$) is followed by a pointwise rectification non-linearity (ReLU)~\cite{ReLU}:
\begin{equation}
\label{eq-out}
\begin{split}
\mathbf{H}^{p} &= \max(0,\mathbf{W}^{p}\mathbf{H}^{p-1} + \mathbf{b}^p)\,,\ p\in\{7,8,9\}\,,\\
\mathbf{Y}^{\phantom{1}} &= \mathbf{W}^{10}\mathbf{H}^{9} + \mathbf{b}^{10}\,. \\
\end{split}
\end{equation}
Parameters of the $p^{th}$ layer are denoted with $(\mathbf{W}^p,\mathbf{b}^p)$. On this step, we do not use any max-pooling. A dropout regularization strategy~\cite{srivastava14a} is applied on all layers.  The network outputs $|\mathcal{C}|+1$ feature planes of dimensions $h^{o}\times w^{o}$, one for each class considered on training, plus background.

%%%%%%%%%%%%%%%%%%%
% Deep Multiple Instance Learning
%%%%%%%%%%%%%%%%%%%
\subsection{Multiple Instance Learning}
\label{dmil}
The network produces one score $s^k_{i,j} = Y^{k}_{i,j}$ for each pixel location
$(i,j)$ from the subsampled image $I$, and for each class
$k\in\mathcal{C}$.  Given that at training time we have only access to
image classification labels, we need a way to aggregate these pixel-level
scores into a single image-level classification score $s^k =
aggreg_{i,j}(s^{k}_{i,j})$, that will then be maximized for the right
class label $k^\star$. Assuming an aggregation procedure $aggreg()$ is chosen, we
interpret image-level class scores as class conditional probabilities by
applying a softmax~\cite{softmax}:
\begin{equation}
p(k|I,\theta) = \frac{e^{s^k}}{\sum_{c\in\mathcal{C}} e^{s^c}}\,,
\end{equation}
where $\theta=\{\mathbf{W}^p,\mathbf{b}^p \ \forall p\}$ represents all the trainable parameters of our architecture.
We then maximize the log-likelihood (with respect to $\theta$), over all the training dataset pairs $(I,k^\star)$:
\begin{equation}
\label{log-likelihood}
\mathcal{L}(\theta) = \sum_{(k^\star,I)}\left[ s^{k^\star} - \log \sum_{c\in\mathcal{C}} e^{s^c} \right]\,.
\end{equation}
Training is achieved with stochastic gradient, backpropagating through the softmax, the aggregation procedure, and up the to first non-frozen layers of our network.

\paragraph{Aggregation}
The aggregation should drive the network towards correct pixel-level
assignments, such that it could perform decently on segmentation tasks.  An
obvious aggregation would be to take the sum over all pixel positions:
\begin{equation}
s^k = \sum_{i,j} s^{k}_{i,j} \quad \forall k\in\mathcal{C}.
\end{equation}
This would however assigns the same weight on all pixels of the image
during the training procedure, even to the ones which do not belong to the
class label assigned to the image. Note that this aggregation method is equivalent as applying a traditional fully-connected classification CNN with a mini-batch. Indeed, each value in the $h^o\times w^o$ output plane corresponds to the output of the CNN fed with a sub-patch centered around the correspond pixel in the input plane.
% In practice, this performs very poorly
% as many pixels are ``background'' (or even from another class) pixels.  
At the other end, one could apply a max pooling aggregation:
\begin{equation}
s^k = \max_{i,j} s^{k}_{i,j} \quad \forall k\in\mathcal{C}.
\end{equation}
This would encourage the model to increase the score of the pixel which is
considered as the most important for image-level classification. In our
experience, this type of approach does not train very well. Note that at
the beginning of the training all pixels might have the same (wrong) score,
but only one (selected by the max) will have its score increased at each
step of the training procedure. It is thus not surprising it takes an
enormous amount of time to the model to converge.

We chose instead a smooth version and convex approximation of the
\emph{max} function, called \emph{Log-Sum-Exp} (LSE)~\cite{boyd2004convex}:
\begin{equation}
s^k = \frac{1}{r}\log\left[\frac{1}{h^o\,w^o}\sum_{i,j}\exp(r\, s^{k}_{i,j})\right]\,.
\end{equation}
The hyper-parameter $r$ controls how smooth one wants the approximation to
be: high $r$ values implies having an effect similar to the max, very low
values will have an effect similar to the score averaging. The advantage of
this aggregation is that pixels having similar scores will have a similar
weight in the training procedure, $r$ controlling this notion of ``similarity''.

\begin{figure*}[!th]
\begin{center}
   \includegraphics[width=\linewidth]{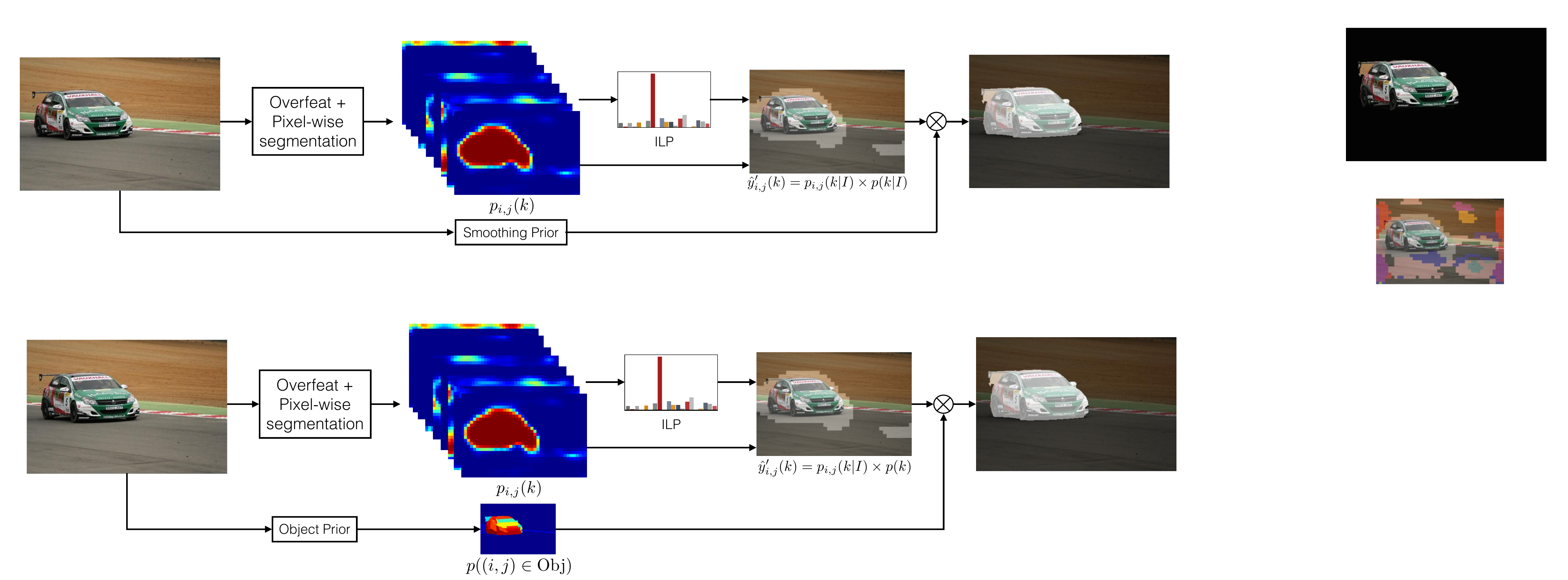}
\end{center}
   \caption{{\bf Inference Pipeline.} The test image is forwarded through the segmentation network to generate a $(|\mathcal{C}|+1)\times h\times w$ output, one plane for each class. The image-level prior is extracted from these planes and the class of each pixel is selected by taking the maximum probability for each pixel. A smoothing prior is also considered to generate a smoother segmentation output.}
\label{fig:test-pipeline}
\end{figure*}

%%%%%%%%%%%%%%%%%%%
% Inference
%%%%%%%%%%%%%%%%%%%
\subsection{Inference}
\label{inference}
At test time, we feed the padded and normalized RGB test image $I$
(of dimension $3\times h\times w$) to our network, where the aggregation
layer has been removed. We thus obtain $|\mathcal{C}|+1$ planes of
pixel-level scores $s^k_{i,j}$ ($1 \leq i \leq h^o$, $1\leq j \leq
w^o$). For convenience (see Section~\ref{sec-adding-segmentation-priors}), we
transform these scores into conditional probabilities $p_{i,j}(k|I)$ using
a softmax over each location $(i,j)$.

Due to the pooling layers in the CNN, the output planes labels correspond
to a sub-sampled version of the input test image. As shown
in~\cite{masci:2013icip,Pinheiro:2014}, one can efficiently retrieve the
label of all pixels of the image using a CNN model, by simply shifting the input image in
both spatial directions, and forwarding it again through the network.

\subsubsection{Adding Segmentation Priors}
\label{sec-adding-segmentation-priors}
Given we do not fine-tune our model on segmentation data, we observed our approach is subject to false positive.
To circumvent this issue, we consider simple post-processing
techniques, namely image-level prior (ILP) and three different smoothing priors (SP), with increasing amount of information. Figure~\ref{fig:test-pipeline} summarizes the pipeline of our approach during inference time.

% - Image-Level Priors
\paragraph{Image-Level Prior}
The model makes inference using local context based on the patch
surrounding a pixel. In order to improve the overall per-pixel accuracy, we
add the global context information of the scene into play. We propose the
use of an image-level prior (ILP)~\cite{ShottonJC08,VezhnevetsB10} based on
the output feature planes. This prior, which is extracted from the trained network, is important to reduce the number of false positives generated by hte model.
As at training time, the probability $p(k|I)$ of each class $k\in\mathcal{C}$ to be present in the scene can be computed by applying the softmax in the LSE score of each label plane. This probability is used as the image-level prior to encourage the likely categories and discourage the unlikely ones. 

The ILP is integrated into the system by multiplying each conditional probability $p_{i,j}(k|I)$ by its class ILP, that is:
\begin{equation}
\hat{y}_{i,j}^{\prime}(k) = p_{i,j}(k|I)\times p(k|I)\,,
\end{equation}
for each location $(i,j)$ and class $k\in\mathcal{C}$.

\paragraph{Smoothing Prior}
\label{sp}
Predicting the class of each pixel independently from its neighbors yields
noisy predictions. In general, objects have smooth boundaries and well
defined shapes, different from the background which tends to be amorphous
regions. At test time we considered three different approaches (of
increasing prior knowledge) to impose local regions with strong boundaries
to be assigned to the same label:

\begin{enumerate}
  \item[(i)] \emph{SP-sppxl} smooths the output using standard
    superpixels. We followed the method proposed by
    \cite{Felzenszwalb:2004}, which largely over-segments a given image
    into a set of disjoint components. Prediction smoothing is achieved by
    simply picking the label that appears the most in each superpixel.
    
    \item[(ii)] \emph{SP-bb} leverages \emph{bounding box candidates} to
    improve the smoothing. We picked the BING algorithm \cite{BingObj2014}
    to generate a set of $10^4$ (possibly overlapping) bounding box
    proposals given an image, each bounding box having a score. These scores are normalized to fit the $[0,1]$ interval.
    Each pixel $(i,j)$ in the image is assigned a score (of belonging to an object) by
    summing the score of all bounding box proposals that contains the pixel. The score at each pixel is then converted into a probability $p((i,j)\in \text{Obj})$ by normalizing the sum by the number of boxes containing the pixel. Label smoothing for each pixel $(i,j)$ is then achieved with:
    \begin{equation*}
      \hat{y}_{i,j}=
      \begin{cases}
        k, & \text{if}\;\; \underset{k\in\mathcal{C}}{\max}\quad \hat{y}^{\prime}_{i,j}(k)\times p((i,j)\in\text{Obj}) > \delta_k\\
        0, & \text{otherwise}
      \end{cases},
    \end{equation*}
    where $\delta_k$ ($0 \leq \delta_k < 1$) is a per-class confidence
    threshold and $\hat{y}_{i,j}=0$ means that background class is assigned
    to the pixel.

  \item[(iii)] \emph{SP-seg} is a smoothing prior which has been trained
    with \emph{class-independent} segmentation labels. We consider the
    Multiscale Combinatorial Grouping (MCG) algorithm~\cite{APBMM2014},
    which generates a serie of overlapping object candidates with a
    corresponding score.  Pixel label smoothing is then achieved in the
    same way as in $\emph{SP-bb}$.
    
\end{enumerate}
The smoothing prior improves our algorithm in two ways: (i) it forces
pixels with low probability of being part of an object to be labeled as
background and (ii) it guarantees local label consistency. While the former
reduces the number of false positives, the latter increases the number of
true positives.  We will see in Section~\ref{experiments} that (as it can
be expected) more complex smoothing priors improves performance accuracy.

%%%%%%%%%%%%%%%%%%%%%%%%%%%%%%%%%%%%%%%%%%%%%%%%%%%%%%%%%%%%%%%
% -- Experiments
%%%%%%%%%%%%%%%%%%%%%%%%%%%%%%%%%%%%%%%%%%%%%%%%%%%%%%%%%%%%%%%
\section{Experiments}
\label{experiments}
Given that our model uses only weak supervision labels (class labels), and is
never trained with segmentation data, we compare our approach with current
state-of-the-art \emph{weakly supervised} segmentation systems. We also
compare it against state-of-the-art \emph{fully supervised} segmentation
systems, to demonstrate that weakly supervised segmentation is a promising
and viable solution.

%%%%%%%%%%%%%%%%%%%
% Datasets
%%%%%%%%%%%%%%%%%%%
\subsection{Datasets}
We considered the Pascal VOC dataset as a benchmark for segmentation. This
dataset includes $20$ different classes, and represents a particular
challenge as an object segmentation task. The objects from these classes
can appear in many different poses, possibly highly occluded, and also
possess a very large intra-class variation. The dataset was only used for
testing purposes, not for training.

We created a large classification training set from the Imagenet dataset
containing images of each of the twenty classes and also an extra class
labeled as \verb+background+ -- set of images in which none of the classes
appear.  We consider all the sub-classes located below each of the twenty
classes in the full Imagenet tree, for a total of around $700,000$
samples. For the background, we chose a subset of Imagenet consisting of a
total of around $60,000$ images not containing any of the twenty
classes\footnote{60K background images might look surprisingly not large, but
we found not easy to pick images where none of the 20 Pascal VOC classes were not present.}. To increase the size of the training
set, jitter (horizontal flip, rotation, scaling, brightness and contrast
modification) was randomly added to each occurrence of an image during the
training procedure. Each image was then normalized for each RGB channel. No other preprocessing was done during training.

%%%%%%%%%%%%%%%%%%%
% Experimental Setup
%%%%%%%%%%%%%%%%%%%
\subsection{Experimental Setup}

\begin{table}[t]%\centering
% \scalebox{0.7}{
\ra{1.3}
\begin{center}
\begin{tabular}{@{}ccccc@{}}
\hline\toprule
{\bf Conv. Layer} & $1$ & $2$ & $3$& $4$ \\
\midrule

{\bf \# channels} & $1024$ & $768$ & $512$ & $21$ \\
{\bf Filter Size} & $3\times3$ & $3\times3$ & $3\times3$ & $3\times3$ \\
{\bf Input Size} & $29\times29$ & $27\times27$ &$ 25\times25$ & $23\times23$ \\

\bottomrule
\hline
\end{tabular}
% }
\end{center}
\caption{{\bf Architecture Design.} Architecture of the segmenter network used in our experiments.}
\label{table:model}
\end{table}

Each training sample consists of a central patch of size $400\times400$
randomly extracted from a deformed image in the training set. If the image
dimensions are smaller than $400\times400$, it is rescaled such that its
smaller dimension is of size 400.

The first layers of our network are extracted (and ``frozen'') from the
public available
\verb+Overfeat+\footnote{\url{http://cilvr.nyu.edu/doku.php?id=software:overfeat:start}}
model. In all our experiments, we use the \emph{slow} Overfeat model, as
described in~\cite{Sermanet:2013vi}. With the $400\times400$ RGB input
image, the \verb+Overfeat+ feature extractor outputs $1024$ feature maps of
dimension $29\times29$. As detailed in Section~\ref{architecture}, these
feature maps are then fed into 4 additional convolutional layers followed
by ReLU non-linearity. A dropout procedure with a rate of 0.5 is applied on
each layer.  The whole network has a total of around 20 million
parameters. Table~\ref{table:model} details the architecture used in our
experiments.

The final convolution layer outputs a $21$ feature maps of dimension
$21\times21$. These feature maps are passed through the aggregation layer
(in the case of LSE, we consider $r=5$), which outputs 21 scores,
one for each class. These scores are then transformed into posterior
probabilities through a \emph{softmax} layer.

Design architecture and hyper-parameters were chosen considering the
validation data of the Pascal VOC 2012 segmentation dataset.
We considered a learning rate $\lambda=0.001$ which decreases by a factor of $0.8$ for every 5 million examples seen by the model. We trained our model using stochastic gradient descent with a batch size of $16$ examples, momentum $0.9$ and weight decay of $0.00005$.

The optimal class confidence thresholds $\delta_k$ for smoothing priors (see
Section~\ref{sp}) were chosen through a grid search. The AP changes in function of the confidence threshold for each class. The different values for the threshold is due to the variability of each class in the training data and how their statistics approach the Pascal VOC images statistics.

Our network takes about a week to train on a Nvidia GeForce Titan GPU
with 6GB of memory. All the experiments were conducted using
Torch7\footnote{\url{http://torch.ch}}.

%%%%%%%%%%%%%%%%%%%
% Experimental Results
%%%%%%%%%%%%%%%%%%%
\begin{table}[!t]%\centering
% \scalebox{0.7}{
\ra{1.3}
\begin{center}
\begin{tabular}{@{}rccc@{}}
\hline\toprule
{\bf Model} & VOC2008 & VOC2009 & VOC2010 \\
\midrule
{\bf MIM}        & $8.11\%$ & $38.27\%$ & $28.43\%$ \\
{\bf GMIM}       & $9.24\%$ & $39.16\%$ & $29.71\%$ \\
{\bf PGC}        & $30.12\%$ & $43.37\%$ & $32.14\%$ \\
{\bf aggreg-max} & $44.31\%$ & $45.46\%$ & $45.88\%$ \\
{\bf aggreg-sum} & $47.54\%$ & $50.01\%$ & $50.11\%$ \\
{\bf aggreg-LSE} & $\mathbf{56.25}\%$ & $\mathbf{57.01}\%$ & $\mathbf{56.12}\%$ \\
\bottomrule
\hline
\end{tabular}
% }
\end{center}
\caption{{\bf Comparison with weakly supervised.} Averaged per-class accuracy of weakly supervised models and ours for different Pascal VOC datasets. We consider three different aggregation layers.}
\label{table:weakly-results}
\end{table}

\begin{table*}[!t]
\begin{center}
\scalebox{0.69}{
\ra{1.3}
\begin{tabular}{@{}rccccccccccccccccccccccc@{}}
\hline\toprule
& \begin{turn}{90}bgnd\end{turn} &\begin{turn}{90} aero \end{turn} &\begin{turn}{90} bike \end{turn} &\begin{turn}{90} bird \end{turn} &\begin{turn}{90} boat \end{turn} &\begin{turn}{90} bottle \end{turn} &\begin{turn}{90} bus \end{turn} &\begin{turn}{90} car \end{turn} &\begin{turn}{90} cat \end{turn} &\begin{turn}{90} chair \end{turn} &\begin{turn}{90} cow \end{turn} &\begin{turn}{90} table \end{turn} &\begin{turn}{90} dog \end{turn} &\begin{turn}{90} horse \end{turn} &\begin{turn}{90} mbike \end{turn} &\begin{turn}{90} person \end{turn} &\begin{turn}{90} plant \end{turn} &\begin{turn}{90} sheep \end{turn} &\begin{turn}{90} sofa \end{turn} &\begin{turn}{90} train \end{turn} &\begin{turn}{90} tv \end{turn} &\begin{turn}{90}  \end{turn} &\begin{turn}{90} mAP\end{turn} \\
\midrule
Fully Sup.\\
{\bf O\textsubscript{2}P}  & $86.1$ & $64.0$ & $27.3$ & $54.1$ & $39.2$ & $48.7$ & $56.6$ & $57.7$ & $52.5$ & $14.2$ & $54.8$ & $29.6$ & $42.2$ & $58.0$ & $54.8$ & $50.2$ & $36.6$ & $58.6$ & $31.6$ & $48.4$ & $38.6$ &  & $47.8$ \\

{\bf DivMBest}  & $85.7$ & $62.7$ & $25.6$ & $46.9$ & $43.0$ & $54.8$ & $58.4$ & $58.6$ & $55.6$ & $14.6$ & $47.5$ & $31.2$ & $44.7$ & $51.0$ & $60.9$ & $53.5$ & $36.6$ & $50.9$ & $30.1$ & $50.2$ & $46.8$ & & $48.1$ \\

{\bf SDS}  & $86.3$ & $63.3$ & $25.7$ & $63.0$ & $39.8$ & $59.2$ & $70.9$ & $61.4$ & $54.9$ & $16.8$ & $45.0$ & $48.2$ & $50.5$ & $51.0$ & $57.7$ & $63.3$ & $31.8$ & $58.7$ & $31.2$ & $55.7$ & $48.5$ & & $51.6$ \\

\midrule
Weak. Sup.\\
{\bf Ours-sppxl} & $74.7$ & $38.8$ & $19.8$ & $27.5$ & $21.7$ & $32.8$ & $40.0$ & $50.1$ & $47.1$ & $7.2$ & $44.8$ & $15.8$ & $49.4$ & $47.3$ & $36.6$ & $36.4$ & $24.3$ & $44.5$ & $21.0$ & $31.5$ & $41.3$ &  & $35.8$ \\
{\bf Ours-bb} & $76.2$ & $42.8$ & $20.9$ & $29.6$ & $25.9$ & $38.5$ & $40.6$ & $51.7$ & $49.0$ & $9.1$ & $43.5$ & $16.2$ & $50.1$ & $46.0$ & $35.8$ & $38.0$ & $22.1$ & $44.5$ & $22.4$ & $30.8$ & $43.0$ &  & $37.0$ \\
{\bf Ours-seg} & $78.7$ & $48.0$ & $21.2$ & $31.1$ & $28.4$ & $35.1$ & $51.4$ & $55.5$ & $52.8$ & $7.8$ & $56.2$ & $19.9$ & $53.8$ & $50.3$ & $40.0$ & $38.6$ & $27.8$ & $51.8$ & $24.7$ & $33.3$ & $46.3$ &  & $40.6$ \\
\bottomrule
\hline
\end{tabular}
}
\end{center}
\caption{{\bf Comparison with fully supervised.} Per class average precision and mean average precision (mAP) on Pascal VOC 2012 segmentation challenge \emph{test set}. We consider different smoothing priors in our model.}
\label{table:fully-results}
\end{table*}

\subsection{Experimental Results}
\paragraph{Compared to weakly supervised models}
We compare the proposed algorithm with three state-of-the-art approaches in
weakly supervised segmentation scenario: (i)~Multi-Image Model
(MIM)~\cite{eth_biwi_00866}, (ii)~a variant, Generalized Multi-Image Model
(GMIM)~\cite{VezhnevetsB12} and (iii)~the most recent Probabilistic
Graphlet Cut (PGC)~\cite{ZhangGXLSJ14,ZhangSLLBC13}.
Note that there are variations in the experimental setup on the experiments. The compared models use Pascal VOC for weak supervision while we use Imagenet. Also, (iii) considers additional labels on the data. 
In our training framework, the Pascal VOC dataset was used only for selecting the thresholds on the class priors.
% The Pascal VOC is not used in our training framework other than selecting the threshold for the priors. 
Our system learns features that are independent of the Pascal VOC data distribution and would a priori yields similar results in other datasets.

Table~\ref{table:weakly-results} reports the results of the three compared models and our approach. In our experiments, we consider the \emph{SP-sppxl} smoothing prior, which does not take into account any segmentation or bounding box information. We consider the three aggregation layers described in Section~\ref{dmil}. This result empirically demonstrates our choice of the \emph{Log-Sum-Exp} layer.

The results for the compared models reported on this table are from Zhang \etal~\cite{ZhangGXLSJ14}. We use the same metric and evaluate on the same datasets (Pascal VOC 2008, 2009 and 2010) as the authors. The metric used, average per-class accuracy, is defined by the ratio of correct classified pixels of each class. We show that our model achieves significantly better results than the previous state-of-the-art weakly supervised algorithms, with an increase from 30\% to 90\% in average per-class accuracy.

\paragraph{Compared to fully supervised models}

In table~\ref{table:fully-results}, we compare the performance of our model
against the best performers in Pascal VOC 2012 segmentation competition:
Second Order Pooling (O\textsubscript{2}P)~\cite{Carreira:2012},
DivMBest~\cite{YadollahpourBS13} and Simultaneous Detection and
Segmentation (SDS)~\cite{hariharan14sds}.  Average precision metric\footnote{$\text{AP}=\frac{TruePositive}{TruePositive + FalsePositive + FalseNegative}$}, as
defined by the Pascal VOC competition, is reported. We show results using all three smoothing priors (as described in~\ref{sp}). The performance of our model increases as we consider more complex priors.

We reach near state-of-the-art performance for several classes (even with
the simplest smoothing prior \emph{SP-sppxl}, which is object and
segmentation agnostic) while some other classes perform worse.  This is not
really surprising, given that the statistics of the images for some classes
(\eg \verb+dog+, \verb+cat+, \verb+cow+) are closer in the two different
datasets than for some other classes (\eg \verb+bird+, \verb+person+). The
results on the specific Pascal VOC challenge could be improved by
``cheating'' and considering training images that are more similar to those
represented on the test data (\eg instead of choosing all \verb+bird+
images from Imagenet, we could have chosen the bird breeds that are similar
to the ones presented on Pascal VOC).

\begin{figure*}[!ht]
 \centering
        \begin{subfigure}[b]{0.16\textwidth}
            \includegraphics[width=\textwidth]{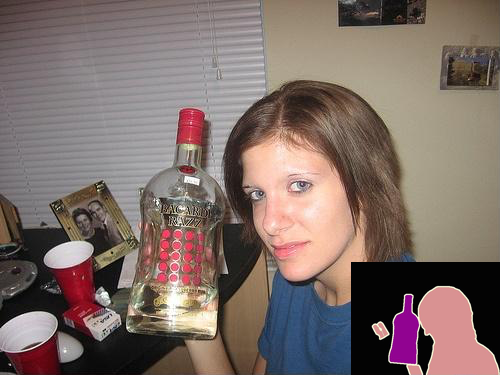}
        \end{subfigure}%
        \hspace{-1mm}
        \begin{subfigure}[b]{0.16\textwidth}
            \includegraphics[width=\textwidth]{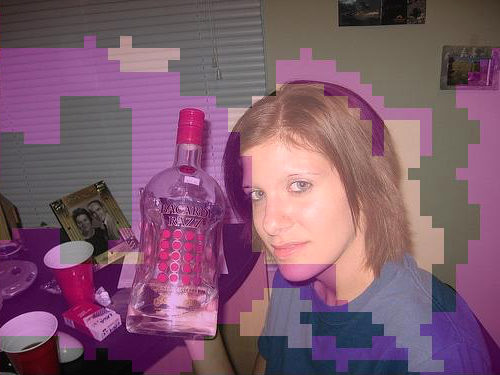}
        \end{subfigure}%
        \hspace{-1mm}
        \begin{subfigure}[b]{0.16\textwidth}
            \includegraphics[width=\textwidth]{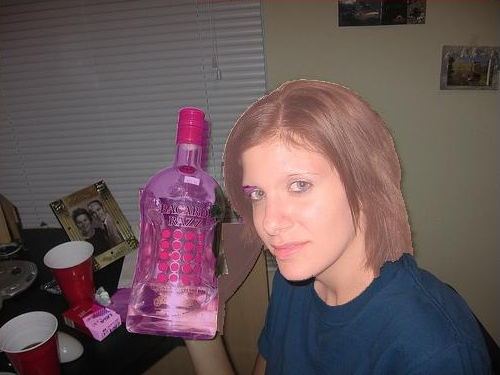}
        \end{subfigure}%
         \hspace{-1mm}
        \begin{subfigure}[b]{0.16\textwidth}
            \includegraphics[width=\textwidth]{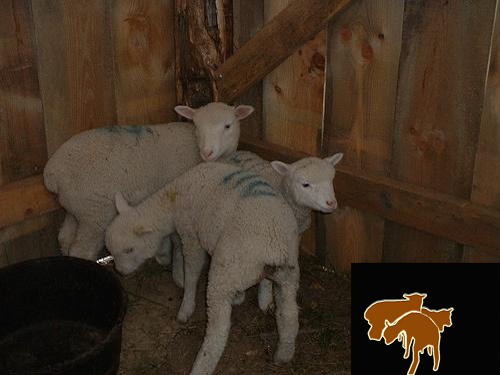}
        \end{subfigure}%
        \begin{subfigure}[b]{0.16\textwidth}
            \includegraphics[width=\textwidth]{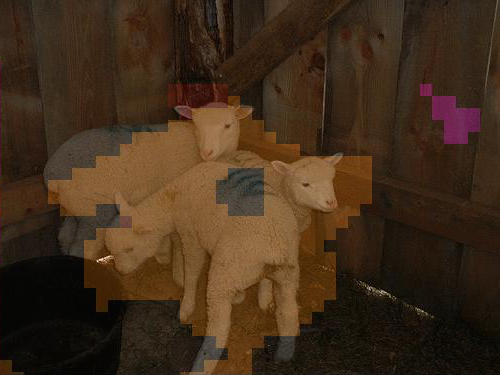}
        \end{subfigure}%
         \hspace{-1mm}
        \begin{subfigure}[b]{0.16\textwidth}
            \includegraphics[width=\textwidth]{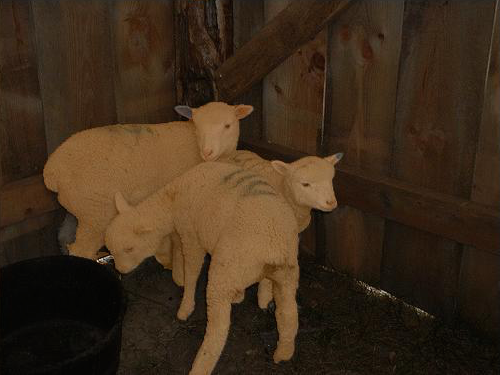}
        \end{subfigure}%
        \\
        \begin{subfigure}[b]{0.16\textwidth}
            \includegraphics[width=\textwidth]{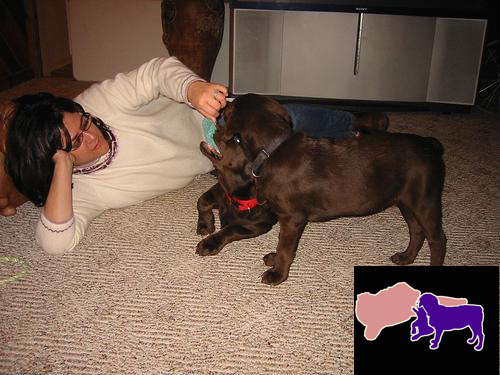}
        \end{subfigure}%
        \hspace{-1mm}
        \begin{subfigure}[b]{0.16\textwidth}
            \includegraphics[width=\textwidth]{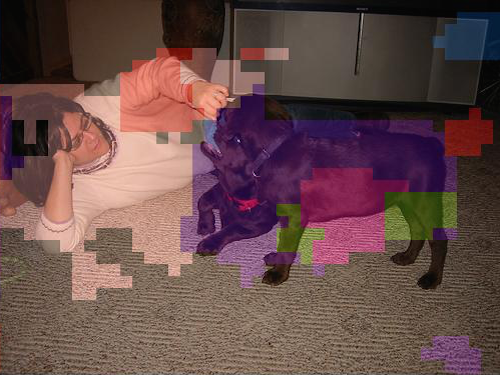}
        \end{subfigure}%
        \hspace{-1mm}
        \begin{subfigure}[b]{0.16\textwidth}
            \includegraphics[width=\textwidth]{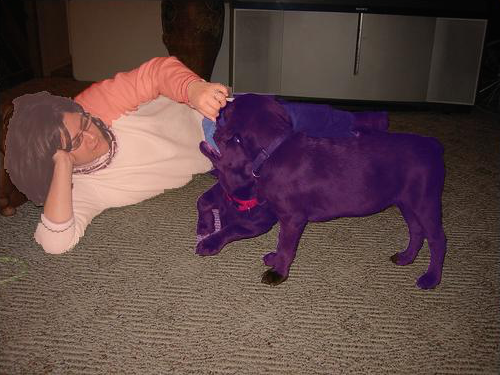}
        \end{subfigure}%
         \hspace{-1mm}
        \begin{subfigure}[b]{0.16\textwidth}
            \includegraphics[width=\textwidth]{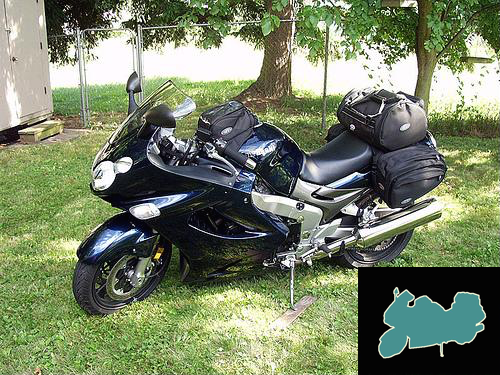}
        \end{subfigure}%
        \begin{subfigure}[b]{0.16\textwidth}
         \includegraphics[width=\textwidth]{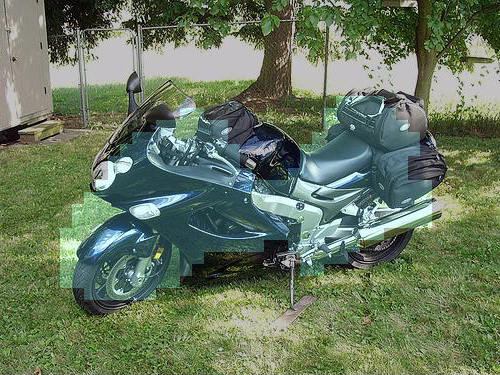}
        \end{subfigure}%
         \hspace{-1mm}
        \begin{subfigure}[b]{0.16\textwidth}
            \includegraphics[width=\textwidth]{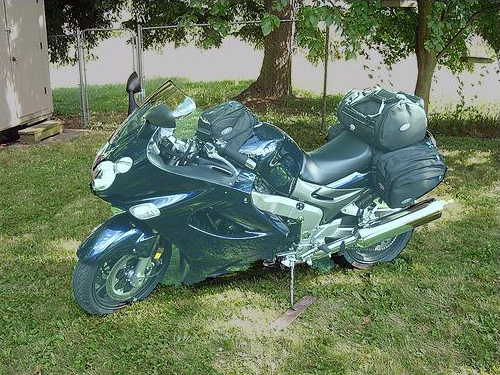}
        \end{subfigure}%
        \\
        \begin{subfigure}[b]{0.16\textwidth}
            \includegraphics[width=\textwidth]{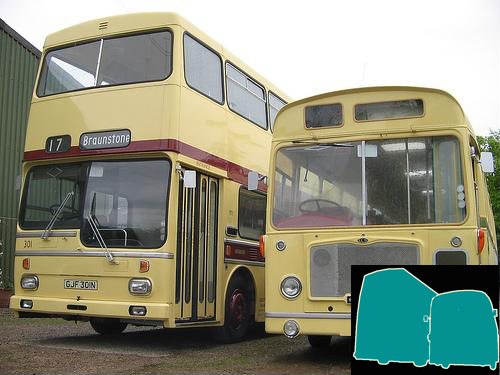}
        \end{subfigure}%
        \hspace{-1mm}
        \begin{subfigure}[b]{0.16\textwidth}
             \includegraphics[width=\textwidth]{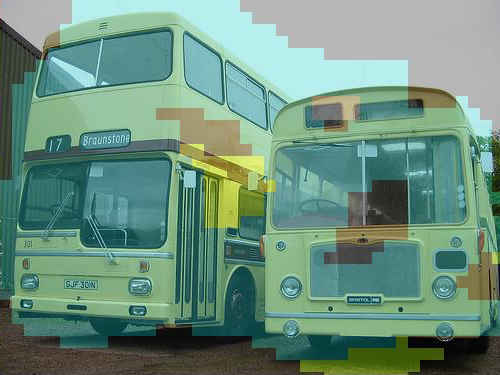}
        \end{subfigure}%
        \hspace{-1mm}
        \begin{subfigure}[b]{0.16\textwidth}
            \includegraphics[width=\textwidth]{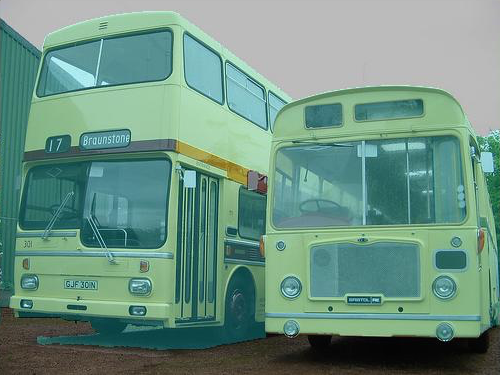}
        \end{subfigure}%
         \hspace{-1mm}
        \begin{subfigure}[b]{0.16\textwidth}
            \includegraphics[width=\textwidth]{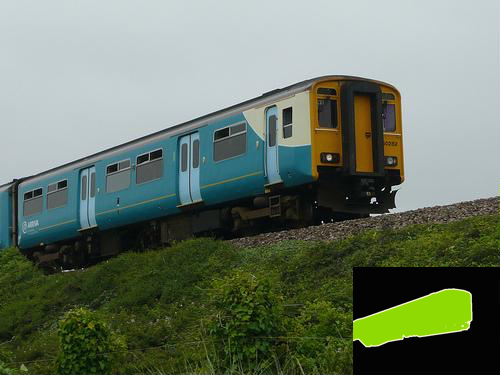}
        \end{subfigure}%
        \begin{subfigure}[b]{0.16\textwidth}
            \includegraphics[width=\textwidth]{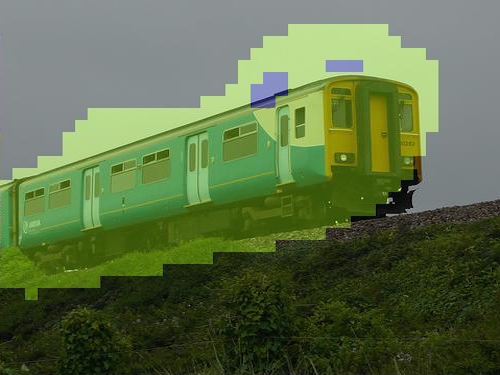}
        \end{subfigure}%
         \hspace{-1mm}
        \begin{subfigure}[b]{0.16\textwidth}
            \includegraphics[width=\textwidth]{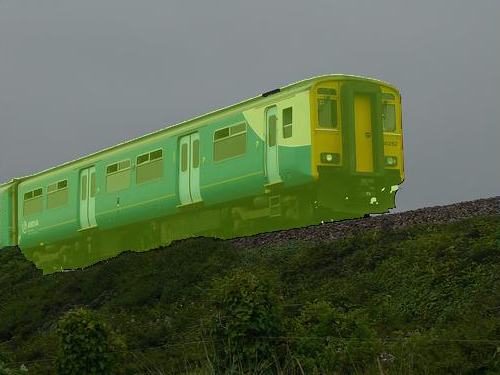}
        \end{subfigure}%
        \\
        \begin{subfigure}[b]{0.16\textwidth}
            \includegraphics[width=\textwidth]{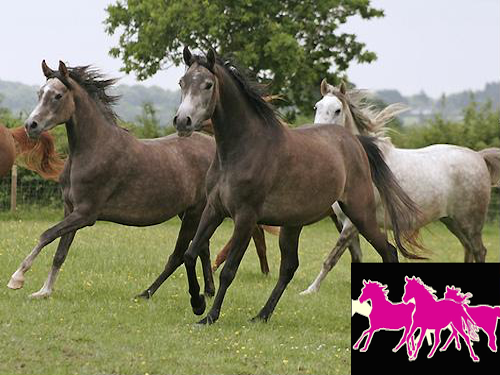}
        \end{subfigure}%
        \hspace{-1mm}
        \begin{subfigure}[b]{0.16\textwidth}
            \includegraphics[width=\textwidth]{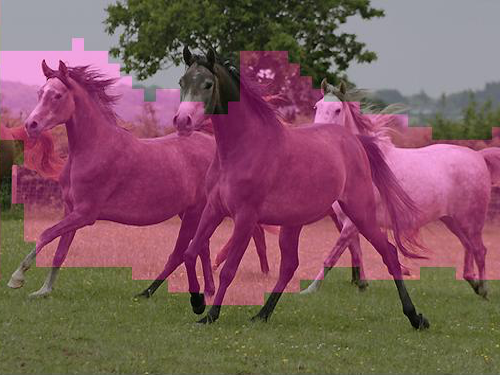}
        \end{subfigure}%
        \hspace{-1mm}
        \begin{subfigure}[b]{0.16\textwidth}
            \includegraphics[width=\textwidth]{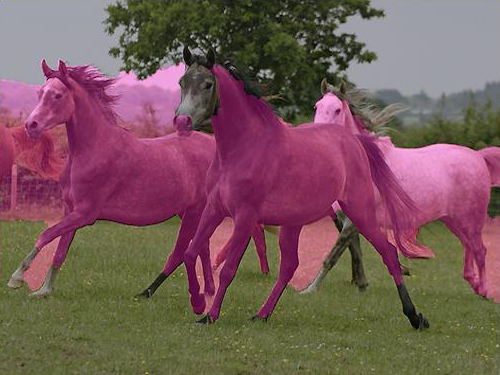}
        \end{subfigure}%
         \hspace{-1mm}
        \begin{subfigure}[b]{0.16\textwidth}
            \includegraphics[width=\textwidth]{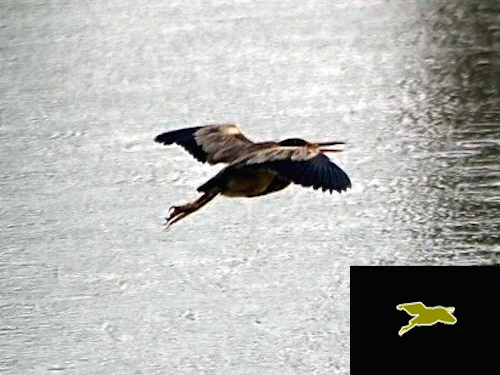}
        \end{subfigure}%
        \begin{subfigure}[b]{0.16\textwidth}
            \includegraphics[width=\textwidth]{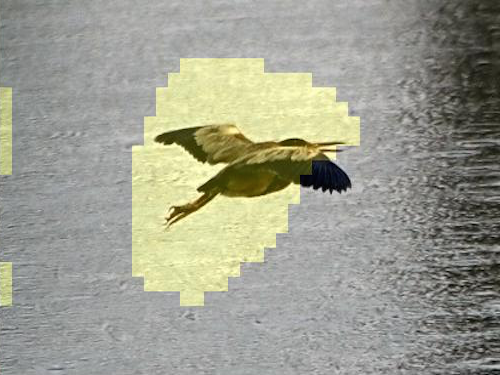}
        \end{subfigure}%
         \hspace{-1mm}
        \begin{subfigure}[b]{0.16\textwidth}
            \includegraphics[width=\textwidth]{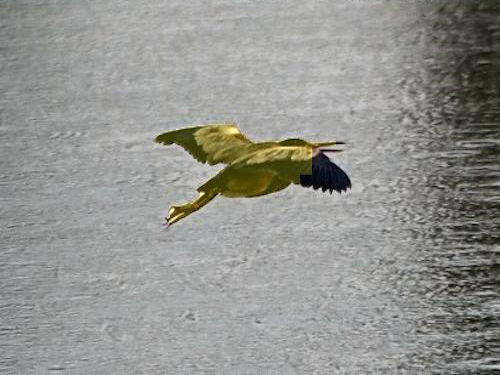}
        \end{subfigure}%
        \\
        \caption{{\bf Inference results}. For each test image (left), we show the output assuming the image-level prior (center) and image-level and \emph{SP-seg} smoothing prior (right).}
        \label{fig:results}
\end{figure*}

\begin{table*}[!h]
\begin{center}
\scalebox{0.65}{
\ra{1.3}
\begin{tabular}{@{}rccccccccccccccccccccccc@{}}
\hline\toprule
& \begin{turn}{90}bgnd\end{turn} &\begin{turn}{90} aero \end{turn} &\begin{turn}{90} bike \end{turn} &\begin{turn}{90} bird \end{turn} &\begin{turn}{90} boat \end{turn} &\begin{turn}{90} bottle \end{turn} &\begin{turn}{90} bus \end{turn} &\begin{turn}{90} car \end{turn} &\begin{turn}{90} cat \end{turn} &\begin{turn}{90} chair \end{turn} &\begin{turn}{90} cow \end{turn} &\begin{turn}{90} table \end{turn} &\begin{turn}{90} dog \end{turn} &\begin{turn}{90} horse \end{turn} &\begin{turn}{90} mbike \end{turn} &\begin{turn}{90} person \end{turn} &\begin{turn}{90} plant \end{turn} &\begin{turn}{90} sheep \end{turn} &\begin{turn}{90} sofa \end{turn} &\begin{turn}{90} train \end{turn} &\begin{turn}{90} tv \end{turn} &\begin{turn}{90}  \end{turn} &\begin{turn}{90} mAP\end{turn} \\
\midrule

{\bf base}  & 37.0 & 10.4 & 12.4 & 10.8 & 5.3 & 5.7 & 25.2 & 21.1 & 25.15 & 4.8 & 21.5 & 8.6 & 29.1 & 25.1 & 23.6 & 25.5 & 12.0 & 28.4 & 8.9 & 22.0 & 11.6 &  & 17.8 \\

{\bf base+ILP}  & 73.2 & 25.4 & 18.2 & 22.7 & 21.5 & 28.6 & 39.5 & 44.7 & 46.6 & 11.9 & 40.4 & 11.8 & 45.6 & 40.1 & 35.5 & 35.2 & 20.8 & 41.7 & 17.0 & 34.7 & 30.4 &  & 32.6 \\

{\bf base+ILP+SP-sppxl}  & 77.2 & 37.3 & 18.4 & 25.4 & 28.2 & 31.9 & 41.6 & 48.1 & 50.7 & 12.7 & 45.7 & 14.6 & 50.9 & 44.1 & 39.2 & 37.9 & 28.3 & 44.0 & 19.6 & 37.6 & 35.0 &  & 36.6 \\

{\bf base+ILP+SP-bb}  & 78.6 & 46.9 & 18.6 & 27.9 & 30.7 & 38.4 & 44.0 & 49.6 & 49.8 & 11.6 & 44.7 & 14.6 & 50.4 & 44.7 & 40.8 & 38.5 & 26.0 & 45.0 & 20.5 & 36.9 & 34.8 &  & 37.8 \\

{\bf base+ILP+SP-seg}  & 79.6 & 50.2 & 21.6 & 40.6 & 34.9 & 40.5 & 45.9 & 51.5 & 60.6 & 12.6 & 51.2 & 11.6 & 56.8 & 52.9 & 44.8 & 42.7 & 31.2 & 55.4 & 21.5 & 38.8 & 36.9 &  & 42.0 \\

\bottomrule
\hline
\end{tabular}
}
\end{center}
\caption{{\bf Effect of priors on segmentation.} Per class average precision on Pascal VOC 2012 \emph{validation set}. We consider the inference with no priors (base), with image-level prior (base+ILP) and different smoothing priors (base+ILP+SP-sppxl, base+ILP+SP-bb, base+ILP+SP-seg).}
\label{table:step-results}
\end{table*}

%%%%%%%%%%%%%%%%%%%
% Effects of Inference Components
%%%%%%%%%%%%%%%%%%%
\paragraph{Effect of Priors}
Table~\ref{table:step-results} shows the average precision of each class on
the Pascal VOC 2012 validation set considering the inference assuming
no prior was used (base), only the image-level prior (base+ILP) and the image-level together with different smoothing priors (base+ILP+SP-sppxl, base+ILP+SP-bb, base+ILP+SP-seg). Figure~\ref{fig:results} illustrates inference in Pascal
VOC images assuming different steps of inference. Priors have a huge
importance to reduce false positives, and smooth predictions.

%%%%%%%%%%%%%%%%%%%%%%%%%%%%%%%%%%%%%%%%%%%%%%%%%%%%%%%%%%%%%%%
% Conclusion
%%%%%%%%%%%%%%%%%%%%%%%%%%%%%%%%%%%%%%%%%%%%%%%%%%%%%%%%%%%%%%%
\section{Conclusion}
\label{conclusion}
We proposed an innovative framework to segment objects with weakly
supervision only. Our algorithm is able to distinguish, at a pixel level,
the differences between different classes, assuming only few simple prior
knowledge about segmentation. This is an interesting result as one might
circumvent the necessity of using the very costly segmentation datasets and
use only image-level annotations.  Our approach surpasses by a large margin previous state-of-the-art models for weakly supervised segmentation. We also achieve competitive performance (at least for several classes) compared to state-of-the-art fully supervised segmentation systems.

%%%%%%%%%%%%%%%%%%%%%%%%%%%%%%%%%%%%%%%%%%%%%%%%%%%%%%%%%%%%%%%
% Acknowledgments
%%%%%%%%%%%%%%%%%%%%%%%%%%%%%%%%%%%%%%%%%%%%%%%%%%%%%%%%%%%%%%%
\newpage
\subsubsection*{Acknowledgments}
The authors thank the reviewers for their useful feedback and comments.
This work was supported by Idiap Research Institute.

%%%%%%%%%%%%%%%%%%%%%%%%%%%%%%%%%%%%%%%%%%%%%%%%%%%%%%%%%%%%%%%
% -- References
%%%%%%%%%%%%%%%%%%%%%%%%%%%%%%%%%%%%%%%%%%%%%%%%%%%%%%%%%%%%%%%
{\small
\bibliographystyle{ieee}
\bibliography{paper-bib}

\begin{thebibliography}{10}\itemsep=-1pt

\bibitem{APBMM2014}
P.~Arbel\'{a}ez, J.~Pont-Tuset, J.~Barron, F.~Marques, and J.~Malik.
\newblock Multiscale combinatorial grouping.
\newblock In {\em IEEE Conference on Computer Vision and Pattern Recognition
  (CVPR)}, 2014.

\bibitem{boyd2004convex}
S.~Boyd and L.~Vandenberghe.
\newblock {\em Convex optimization}.
\newblock Cambridge University Press, 2004.

\bibitem{softmax}
J.~S. Bridle.
\newblock Probabilistic interpretation of feedforward classification network
  outputs, with relationships to statistical pattern recognition.
\newblock {\em Neurocomputing: Algorithms, Architectures and Applications},
  1990.

\bibitem{Carreira:2012}
J.~Carreira, R.~Caseiro, J.~Batista, and C.~Sminchisescu.
\newblock Semantic segmentation with second-order pooling.
\newblock In {\em European Conference on Computer Vision (ECCV)}, 2012.

\bibitem{BingObj2014}
M.~Cheng, Z.~Zhang, W.~Lin, and P.~H.~S. Torr.
\newblock {BING}: Binarized normed gradients for objectness estimation at
  300fps.
\newblock In {\em IEEE Conference on Computer Vision and Pattern Recognition
  (CVPR)}, 2014.

\bibitem{imagenet_cvpr09}
J.~Deng, W.~Dong, R.~Socher, L.~Li, K.~Li, and L.~Fei-Fei.
\newblock Imagenet: A large-scale hierarchical image database.
\newblock In {\em IEEE Conference on Computer Vision and Pattern Recognition
  (CVPR)}, 2009.

\bibitem{Everingham10}
M.~Everingham, L.~V. Gool, C.~K.~I. Williams, J.~Winn, and A.~Zisserman.
\newblock The pascal visual object classes ({VOC}) challenge.
\newblock {\em International Journal of Computer Vision}, 2010.

\bibitem{Farabet:2012}
C.~Farabet, C.~Couprie, L.~Najman, and Y.~LeCun.
\newblock Learning hierarchical features for scene labeling.
\newblock {\em IEEE Transactions on Pattern Analysis and Machine Intelligence
  (PAMI)}, 2013.

\bibitem{Felzenszwalb:2004}
P.~F. Felzenszwalb and D.~P. Huttenlocher.
\newblock Efficient graph-based image segmentation.
\newblock {\em International Journal of Computer Vision (IJCV)}, 2004.

\bibitem{girshick2014rcnn}
R.~Girshick, J.~Donahue, T.~Darrell, and J.~Malik.
\newblock Rich feature hierarchies for accurate object detection and semantic
  segmentation.
\newblock In {\em IEEE Conference on Computer Vision and Pattern Recognition
  (CVPR)}, 2014.

\bibitem{hariharan14sds}
B.~Hariharan, P.~Arbel\'{a}ez, R.~Girshick, and J.~Malik.
\newblock Simultaneous detection and segmentation.
\newblock In {\em Proceedings of the European Conference on Computer Vision
  ({ECCV})}, 2014.

\bibitem{NIPS2012_0534}
A.~Krizhevsky, I.~Sutskever, and G.~Hinton.
\newblock Imagenet classification with deep convolutional neural networks.
\newblock In {\em Advances in Neural Information Processing Systems (NIPS)}.
  2012.

\bibitem{lecun1998gradient}
Y.~LeCun, L.~Bottou, Y.~Bengio, and P.~Haffner.
\newblock Gradient-based learning applied to document recognition.
\newblock {\em Proceedings of the IEEE}, 1998.

\bibitem{Maron98}
O.~Maron and T.~Lozano-P\'{e}rez.
\newblock A framework for multiple instance learning.
\newblock In {\em Advances in Neural Information Processing Systems (NIPS)},
  1998.

\bibitem{masci:2013icip}
J.~Masci, A.~Giusti, D.~C. Ciresan, G.~Fricout, and J.~Schmidhuber.
\newblock A fast learning algorithm for image segmentation with max-pooling
  convolutional networks.
\newblock In {\em International Conference on Image Processing (ICIP13)}, 2013.

\bibitem{ReLU}
V.~Nair and G.~E. Hinton.
\newblock Rectified linear units improve restricted boltzmann machines.
\newblock In {\em International Conference on Machine Learning (ICML)}, 2010.

\bibitem{Oquab14}
M.~Oquab, L.~Bottou, I.~Laptev, and J.~Sivic.
\newblock Learning and transferring mid-level image representations using
  convolutional neural networks.
\newblock In {\em IEEE Conference on Computer Vision and Pattern Recognition
  (CVPR)}, 2014.

\bibitem{Oquab13}
M.~Oquab, L.~Bottou, I.~Laptev, and J.~Sivic.
\newblock Weakly supervised object recognition with convolutional neural
  networks.
\newblock Technical report, INRIA, 2014.

\bibitem{Pinheiro:2014}
P.~H.~O. Pinheiro and R.~Collobert.
\newblock Recurrent convolutional neural networks for scene labeling.
\newblock In {\em International Conference on Machine Learning (ICML)}, 2014.

\bibitem{RazavianASC14}
A.~S. Razavian, H.~Azizpour, J.~Sullivan, and S.~Carlsson.
\newblock {CNN} features off-the-shelf: an astounding baseline for recognition.
\newblock {\em CoRR}, 2014.

\bibitem{Sermanet:2013vi}
P.~Sermanet, D.~Eigen, X.~Zhang, M.~Mathieu, R.~Fergus, and Y.~LeCun.
\newblock {OverFeat: Integrated Recognition, Localization and Detection using
  Convolutional Networks}.
\newblock In {\em International Conference on Learning Representations (ICLR)},
  2014.

\bibitem{ShottonJC08}
J.~Shotton, M.~Johnson, and R.~Cipolla.
\newblock Semantic texton forests for image categorization and segmentation.
\newblock In {\em IEEE Conference on Computer Vision and Pattern Recognition
  (CVPR)}, 2008.

\bibitem{srivastava14a}
N.~Srivastava, G.~Hinton, A.~Krizhevsky, I.~Sutskever, and R.~Salakhutdinov.
\newblock Dropout: A simple way to prevent neural networks from overfitting.
\newblock {\em Journal of Machine Learning Research}, 2014.

\bibitem{GoogLenet}
C.~Szegedy, W.~Liu, Y.~Jia, P.~Sermanet, S.~Reed, D.~Anguelov, D.~Erhan,
  V.~Vanhoucke, and A.~Rabinovich.
\newblock {Going Deeper with Convolutions}.
\newblock 2014.

\bibitem{VezhnevetsB10}
A.~Vezhnevets and J.~M. Buhmann.
\newblock Towards weakly supervised semantic segmentation by means of multiple
  instance and multitask learning.
\newblock In {\em IEEE Conference on Computer Vision and Pattern Recognition
  (CVPR)}, 2010.

\bibitem{eth_biwi_00866}
A.~Vezhnevets, V.~Ferrari, , and J.~Buhmann.
\newblock Weakly supervised semantic segmentation with a multi-image model.
\newblock In {\em Proceedings of the International Conference on Computer
  Vision (ICCV)}, 2011.

\bibitem{VezhnevetsB12}
A.~Vezhnevets, V.~Ferrari, and J.~M. Buhmann.
\newblock Weakly supervised structured output learning for semantic
  segmentation.
\newblock In {\em IEEE Conference on Computer Vision and Pattern Recognition
  (CVPR)}, 2012.

\bibitem{YadollahpourBS13}
P.~Yadollahpour, D.~Batra, and G.~Shakhnarovich.
\newblock Discriminative re-ranking of diverse segmentations.
\newblock In {\em IEEE Conference on Computer Vision and Pattern Recognition
  (CVPR)}, 2013.

\bibitem{ZeilerECCV14}
M.~Zeiler and R.~Fergus.
\newblock Visualizing and understanding convolutional networks.
\newblock In {\em European Conference on Computer Vision (ECCV)}, 2014.

\bibitem{ZhangGXLSJ14}
L.~Zhang, Y.~Gao, Y.~Xia, K.~Lu, J.~Shen, and R.~Ji.
\newblock Representative discovery of structure cues for weakly-supervised
  image segmentation.
\newblock {\em {IEEE} Transactions on Multimedia}, 2014.

\bibitem{ZhangSLLBC13}
L.~Zhang, M.~Song, Z.~Liu, X.~Liu, J.~Bu, and C.~Chen.
\newblock Probabilistic graphlet cut: Exploiting spatial structure cue for
  weakly supervised image segmentation.
\newblock In {\em IEEE Conference on Computer Vision and Pattern Recognition
  (CVPR)}, 2013.

\end{thebibliography}
}

\end{document}